\def\year{2022}\relax
\documentclass[letterpaper]{article} 
\usepackage{aaai22}  
\usepackage{times}  
\usepackage{helvet}  
\usepackage{courier}  
\usepackage[hyphens]{url}  
\usepackage{graphicx} 
\usepackage{natbib}  
\usepackage{caption} 
\DeclareCaptionStyle{ruled}{labelfont=normalfont,labelsep=colon,strut=off} 
\usepackage{algorithm}
\usepackage{algorithmic}
\usepackage{multirow}
\usepackage{makecell}
\usepackage{color}
\usepackage{bm,bbm}
\usepackage{amsmath}
\usepackage{amsfonts}
\newtheorem{theorem}{Theorem}

\usepackage{color,xcolor}
\definecolor{Light}{gray}{0.8}
\usepackage{colortbl}
\usepackage{booktabs}
\usepackage{multicol}
\usepackage{newfloat}
\usepackage{listings}
\floatstyle{ruled}
\newfloat{listing}{tb}{lst}{}
\floatname{listing}{Listing}
\title{Scaled ReLU Matters for Training Vision Transformers}

\newcommand*\samethanks[1][\value{footnote}]{\footnotemark[#1]}
\author {
   Pichao Wang\thanks{The first two authors contribute equally.},
   Xue Wang\samethanks,
   Hao Luo,
   Jingkai Zhou,
   Zhipeng Zhou,
   Fan Wang,
   Hao Li,
   Rong Jin
}
\affiliations {
    Alibaba Group\\
    \texttt{\{pichao.wang,xue.w\}@alibaba-inc.com} \\
  \texttt{\{michuan.lh,zhoujingkai.zjk,yuer.zzp,fan.w,lihao.lh,jinrong.jr\}@alibaba-inc.com} 
}

\usepackage{bibentry}

\begin{document}

\maketitle

\begin{abstract}
Vision transformers (ViTs) have been an alternative design paradigm to convolutional neural networks (CNNs). However, the training of ViTs is much harder than CNNs, as it is sensitive to the training parameters, such as learning rate, optimizer and warmup epoch. The reasons for training difficulty are empirically analysed in ~\cite{xiao2021early},  and the authors conjecture that the issue lies with the \textit{patchify-stem} of ViT models and propose that early convolutions help transformers see better. In this paper, we further investigate this problem and extend the above conclusion: only early convolutions do not help for stable training, but the scaled ReLU operation in the \textit{convolutional stem} (\textit{conv-stem}) matters. We verify, both theoretically and empirically, that scaled ReLU in \textit{conv-stem} not only improves training stabilization, but also increases the diversity of patch tokens, thus boosting peak performance with a large margin via adding few parameters and flops. In addition, extensive experiments are conducted to demonstrate that previous ViTs are far from being well trained, further showing that ViTs have great potential to be a better substitute of CNNs.
\end{abstract}

\section{Introduction}

\noindent Visual recognition has been dominated by convolutional neural networks (CNNs)~\cite{he2016deep,howard2017mobilenets,zhang2018shufflenet,tan2019efficientnet,li2021involution,zhou2021decoupled} for years, which effectively impose spatial locality and translation equivalence. Recently the prevailing vision transformers (ViTs) are regarded as an alternative design paradigm,  which target to replace the inductive bias towards local processing inherent in CNNs with global self-attention~\cite{dosovitskiy2020image,touvron2020training,wang2021pyramid,fan2021multiscale}. 

Despite the appealing potential of ViTs for complete data-driven training, the lack of convolution-like inductive bias also challenges the training of ViTs. Compared with CNNs, ViTs are sensitive to the choice of optimizer, data augmentation, learning rate, training schedule length and warmup epoch~\cite{touvron2020training,touvron2021going,chen2021vision,xiao2021early}. The reasons for training difficulty are empirically analysed in ~\cite{xiao2021early},  and the authors conjecture that the issue lies with the \textit{patchify stem} of ViT models and propose that early convolutions help transformers see better. Recent works~\cite{graham2021levit,guo2021cmt,yuan2021volo} also introduce the \textit{conv-stem} to improve the robustness of training vision transformer, but they lack the deep analysis why such \textit{conv-stem} works. 

In this paper, we theoretically and empirically verify that scaled ReLU in the 
\textit{conv-stem} matters for the robust ViTs training. Specifically, scaled ReLU not only improves the training stabilization, but also increases the diversity of patch tokens, thus boosting the final recognition performances by a large margin. In addition, extensive experiments are conducted to further unveil the effects of \textit{conv-stem} and the following interesting observations are made: firstly, after adding \textit{conv-stem} to the ViTs, the SAM optimizer~\cite{foret2020sharpness} is no longer powerful as reported in~\cite{chen2021vision}; secondly, with \textit{conv-stem}, the supervised ViTs~\cite{touvron2020training} are better than its corresponding self-supervised trained models~\cite{caron2021emerging} plus supervised finetuning on Imagenet-$1$k; thirdly, using \textit{conv-stem} the better trained ViTs improve the performance of downstream tasks. All of these observations reflect that previous ViTs are far from being well trained and ViTs may become 
a better substitute for CNNs.

\section{Related Work}

\textbf{Convolutional neural networks (CNNs)}. Since the breakthrough performance on ImageNet via AlexNet~\cite{krizhevsky2012imagenet}, CNNs have become a dominant architecture in computer vision field. Following the primary design rule of stacking low-to-high convolutions in series by going deeper, many popular architectures are proposed, such as VGG~\cite{simonyan2014very}, GoogleNet~\cite{szegedy2015going} and ResNet~\cite{he2016deep}. To further exploit the capacity of visual representation, many innovations have been proposed, such as ResNeXt~\cite{xie2017aggregated}, SENet~\cite{hu2018squeeze}, EfficientNet~\cite{tan2019efficientnet} and NFNet~\cite{brock2021high}. For most of these CNNs, \textit{Conv+BN+ReLU} becomes a standard block. In this paper, we investigate this basic block for training vision transformers as a lightweight stem.  

\textbf{Vision Transformers (ViTs)}. Since Dosovitskiy et al.~\cite{dosovitskiy2020image} first successfully applies transformer for image classification by dividing the images into non-overlapping patches, many ViT variants are proposed~\cite{wang2021pyramid,han2021transformer,chen2021psvit,ranftl2021vision,liu2021swin,chen2021crossvit,zhang2021multi,xie2021so,zhang2021aggregating,jonnalagadda2021foveater,wang2021not,fang2021msg,huang2021shuffle,gao2021container,rao2021dynamicvit,yu2021glance,zhou2021refiner,el2021xcit,wang2021crossformer,xu2021evo}.  In this section, we mainly review several closely related works for training ViTs.  Specifically, DeiT~\cite{touvron2020training} adopts several training techniques (e.g. truncated normal initialization, strong data augmentation and smaller weight decay) and uses distillation to extend ViT to a data-efficient version;  T2T ViT~\cite{yuan2021tokens}, CeiT~\cite{yuan2021incorporating}, and CvT~\cite{wu2021cvt} try to deal with the rigid patch division by introducing convolution operation for patch sequence generation to facilitate the training;  DeepViT~\cite{zhou2021deepvit}, CaiT~\cite{touvron2021going}, and PatchViT~\cite{gong2021improve} investigate the unstable training problem, and propose the re-attention, re-scale and anti-over-smoothing techniques respectively for stable training; to accelerate the convergence of training, ConViT~\cite{d2021convit}, PiT~\cite{heo2021rethinking}, CeiT~\cite{yuan2021incorporating}, LocalViT~\cite{li2021localvit} and Visformer~\cite{Zhengsu21} introduce convolutional bias to speedup the training;  LV-ViT~\cite{jiang2021token} adopts several techniques including MixToken and Token Labeling for better training and feature generation;  the SAM optimizer~\cite{foret2020sharpness} is adopted in~\cite{chen2021vision} to better train ViTs without strong data augmentation; KVT~\cite{wang2021kvt} introduces the $k$-NN attention to filters out irrelevant tokens to speedup the training; \textit{conv-stem} is adopted in several works~\cite{graham2021levit,xiao2021early,guo2021cmt,yuan2021volo} to improve the robustness of training ViTs. In this paper, we investigate the training of ViTs by using the \textit{conv-stem} and demonstrate several properties of \textit{conv-stem} in the context of vision transformers, both theoretically and empirically.

\section{Vision Transformer Architectures}
In this section, we first review the vision transformer, namely ViT ~\cite{dosovitskiy2020image}, and then describe the \textit{conv-stem} used in our work.

\textbf{ViT}. ViT first divides an input image into non-overlapping $p$x$p$ patches and linearly projects each patch to a $d$-dimensional feature vector using a learned weight matrix. The typical patch and image size are $p$ = 16 and 224x224, respectively. The patch embeddings together with added positional embeddings and a concatenated classification token are fed into a standard transformer encoder~\cite{vaswani2017attention} followed by a classification head. Similar as~\cite{xiao2021early}, we name the portion of ViT before the transformer blocks as \textit{ViT-stem}, and call the linear projection (stride-$p$, $p$x$p$ kernel) as \textit{patchify-stem}. 

\textbf{Conv-stem.} Unless otherwise specified, we adopt the \textit{conv-stem} from VOLO~\cite{yuan2021volo}. The full \textit{conv-stem} consists of 3Conv+3BN+3ReLU+1Proj blocks, and the kernel sizes and strides are (7,3,3,8) and (2,1,1,8), respectively. The detailed configurations are shown in Algorithm 1 of supplemental material. The parameters and FLOPs of \textit{conv-stem} are slightly larger than \textit{patchify-stem}. For example, the parameters of DeiT-Small increase from 22M to 23M, but the increase is very small as the kernel size in last linear projection layer decreases from 16*16 in \textit{patchify-stem}  to 8*8 in \textit{conv-stem}. The reason why we adopt the VOLO \textit{conv-stem} rather than that in~\cite{xiao2021early} is that we want to keep the layers of encoders the same as in ViT, but not to remove one encoder layer as in~\cite{xiao2021early}.

\textbf{ViT$_{p}$ and ViT$_{c}$}. To make easy comparisons, the original ViT model using \textit{patchify-stem} is called ViT$_{p}$. To form a ViT model with a \textit{conv-stem}, we simply replace the \textit{pathify-stem} with \textit{conv-stem}, leaving all the other unchanged, and we call this ViT as ViT$_{c}$. In the following sections, we theoretically and empirically verify that ViT$_{c}$ is better than ViT$_{p}$ in stabilizing training and diversifying the patch tokens, due to the scaled ReLU structure.

\section{Scaled ReLU Structure}
In this section, we first introduce the Scaled ReLU structure and then analyze how scaled ReLU stabilizes training and enhances the token diversification respectively.

For any input $x$, we defined the scaled ReLU structure with scaling parameters $\alpha,\beta$, $ReLU_{\alpha,\beta}(\cdot)$ for shorthand, as follow:
\begin{align}
   ReLU_{\alpha,\beta}(x) =  \beta\max\left\{x+\alpha,0\right\}.\notag
\end{align}
The scaled ReLU structure can be achieved by combining ReLU with normalization layers, such as Batchnorm or Layernorm that contain trainable scaling parameters, and one can view the Batchnorm + ReLU in the {\it conv-stem} as a variant of the scaled ReLU. Intuitively, the ReLU layer may cut out 
part of input data and make the data focus on a smaller range. It is necessary to scale it up to a similar data range as of its input, which helps stabilize training as well as maintain promising expression power. For simplicity, we will focus on the scaled ReLU in this paper and our analysis could be extended to the case with commonly used normalization layers.

\subsection{Training stabilization}
Let's assume $X_{i,c}\in\mathbbm{R}^{HW}$ be the output of channel $c$ in the CNN layer from the last \textit{conv-stem} block for $i$-th sample, where $H$ and $W$ are height and width. Based on the definition of the Batchnorm, the output $X_{i,c}^{out}$ of the last
\textit{conv-stem} is
\begin{align}
    X_{i,c}^{out} &= ReLU\left(\frac{X_{i,c}-\mu_c e}{\sqrt{\sum_{i=1}^B\|X_{i,c}-\mu_c e\|^2}}\beta_c+\alpha_c e\right)\notag\\
    &=ReLU_{\frac{\alpha_c}{\beta_c},\beta_c}\left(\frac{X_{i,c}-\mu_c e}{\sqrt{\sum_{i=1}^B\|X_{i,c}-\mu_c e\|^2}}\right)\notag\\
    &=ReLU_{\frac{\alpha_c}{\beta_c},\beta_c}(\tilde{X}_{i,c}),\label{eq:0_1}
\end{align}
where $\tilde{X}_{i,c} = \frac{X_{i,c}-\mu_c e}{\sqrt{\sum_{i=1}^B\|X_{i,c}-\mu_c e\|^2}}$, $\mu_c$ is the mean of $X_{i,c}$ within a batch and $B$ is the batch size. Next, we concatenate  $X_{i,c}^{out}$ over channel as $X_{i}^{out}$ and reshape it to $X_{i}^{in}\in\mathbbm{R}^{B\times n\times d}$, where $n$ is the token (patch) length and $d$ is the embedding dimension. Finally, we compute $Q_i,K_i,V_i$ as follow:


\begin{align}
    \begin{bmatrix}
    Q_i&
    K_i&
    V_i
    \end{bmatrix} = X_{i}^{in}\begin{bmatrix}
    W_Q&
    W_K&
    W_V
    \end{bmatrix}\doteq X_{i}^{in} W_{trans}\notag
\end{align}
and start to run the self attention. 

To illustrate how the scaled ReLU can stabilize training, we consider a special case which we freeze all parameters except the scaling parameters $\alpha_c,\beta_c$ for $c=1,2,...,C$ in the last batchnorm layer, and $W_Q$, $W_K$ and $W_V$ in the first transformer block. Note that $Q$, $K$ and $V$ are computed by the production of $X^{in}$ and $W_{trans}$. In order to maintain the same magnitude of $Q$, $K$ and $V$, $W_{trans}$ will be closer to $0$ if $X^{in}$ is scaled with larger $\alpha_c$ and $\beta_c$ parameters. In other words, the Scaled ReLU may give the $W_{trans}$ an implicit regularization with respect to its scaling parameters. The result is summarized in the following Theorem \ref{thm:1}.

\begin{theorem}\label{thm:1}
Let $\alpha_c$, $\beta_c$ be the parameters in scaled ReLU structures in the last \textit{conv-stem} block with $c=1,2,...,C$ and $W_{trans}\doteq \begin{bmatrix}
    W_Q&
    W_K&
    W_V
    \end{bmatrix}$ be the attention parameters in the first transformer block. If we freeze all other parameters and introduce the $l_2$ weight decay in the optimizer, then the optimization problem is equivalent to the weighted $l_1$ penalized learning on $W_{trans}$. Moreover, let $W_{trans,c}$ be the parameters associated with channel $c$ and the penalty weights corresponding to $W_{trans,c}$ are proportional to $\sqrt{\beta_c^2+\alpha_c^2}$.
\end{theorem}

The theorem shows an implicit $l_1$ regularization on attention weights from the scaled ReLU structure. In the modern high-dimensional statistics, it is well known that $l_1$ penalized learning introduces significantly less model bias (e.g., exponentially better dimensionality efficiency  shown in \citealt{loh2015regularized}). Moreover, the regularization strength that is on the order of  $\mathcal{O}(\sqrt{\alpha_c^2+\beta_c^2})$ differs from channel to channel and changes over time adaptively. For the channel with larger magnitude in $\alpha_c$ and/or $\beta_c$, the scaled token has higher divergence. In order to make the training processing more stable, the updates for the corresponding parameters in $W_{trans}$ need also be more careful (using larger penalties). It distinguishes the scaled ReLU structure from directly using $l_1$ weights decay in the optimizer directly.

{\it Proof of Theorem \ref{thm:1}}\footnote{
The similar analysis procedure for implicit regularization are also presented in \cite{ergen2021demystifying,neyshabur2014search,savarese2019infinite}.\\}
We denote the loss function as follow:
\begin{align}
    \min &\frac{1}{n}\sum_{i=1}^n KL\left(f(\{ReLU_{\frac{\alpha_c}{\beta_c},\beta_c}(\tilde{X}_{i,c})\},W_{trans}),y_i\right)\notag\\
    &\quad\quad+\lambda \left(\sum_{c=1}^C(\alpha_c^2+\beta_c^2)+\|W_{trans}\|_F^2\right),\notag
\end{align}
where $KL(\cdot)$ is the KL-divergence, $y_i$ is the label for $i$-th sample, $f(\cdot)$ denotes prediction function, $\lambda$ is a positive constant for $l_2$ weight decays and  $\{ReLU_{\alpha_c,\beta_c}(\tilde{X}_{i,c})\}$ is the set of $ReLU_{\alpha_c,\beta_c}(\tilde{X}_{i,c})$ over all channels. Without loss of generality, we can find a function $g$ to rewrite $f$ function as:
\begin{align}
&f\left(\left\{ReLU_{\frac{\alpha_c}{\beta_c},\beta_c}(\tilde{X}_{i,c})\right\},W_{trans}\right)\notag\\
&\quad= g\left(\left\{ReLU_{\frac{\alpha_c}{\beta_c},\beta_c}(\tilde{X}_{i,c})W_{trans,c}\right\}\right),\notag
\end{align}
where we rearrange $W_{trans,c}$ to match the dimensions of the \textit{conv-stem} (i.e., $C\times HW$ instead of $n\times d$). 
Next, we can re-scale the parameters with $\eta_c>0$ as follow:
\begin{align}
    \tilde{\beta}_c = \eta_c\beta_c\notag, 
    \tilde{\alpha}_c = \eta_c\alpha_c\notag, 
    \tilde{W}_{trans,c} = \eta_c^{-1}W_{trans,c}\notag,
\end{align}
and it implies
\begin{align}
    &g\left(\left\{ReLU_{\frac{\alpha_c}{\beta_c},\beta_c}(\tilde{X}_{i,c})W_{trans,c}\right\}\right)\notag\\
    &\quad= g\left(\left\{ReLU_{\frac{\tilde{\alpha}_c}{\tilde{\beta}_c},\tilde{\beta}_c}(\tilde{X}_{i,c})\tilde{W}_{trans,c}\right\}\right).\notag
\end{align}
Moreover, using the fact that $(a^2+b^2)+c^2\ge 2|c|\sqrt{a^2+b^2}$ one can verify
\begin{align}
&\sum_{c=1}^C(\tilde{\alpha}_c^2+\tilde{\beta}_c^2)+\|\tilde{{W}}_{trans}\|_F^2\notag\\
=&\sum_{c=1}^C\tilde{\alpha}_c^2+\tilde{\beta}_c^2+\|\tilde{{W}}_{trans,c}\|^2\notag\\
\ge& 2\sum_{c=1}^C\|\eta_c^{-1}W_{trans,c}\|_{1}\sqrt{\frac{\eta_c^2\alpha_c^2+\eta_c^2\beta_c^2}{HW}}\label{eq:1_1}\\
=& \frac{2}{\sqrt{HW}}\sum_{c=1}^C\|W_{trans,c}\|_{1}\sqrt{\alpha_c^2+\beta_c^2}\label{eq:2},
\end{align}
where the equality \eqref{eq:1_1} holds when 
\begin{align}
    \eta_c = \sqrt{\frac{\|W_{trans,c}\|_1}{\alpha_c^2+\beta_c^2}},\quad c= 1,2,...,C.
\end{align}
Therefore the right hand-size of \eqref{eq:2} becomes the $l_1$ penalties over the $W_{trans,c}$ with weights $\sqrt{\alpha_c^2+\beta_c^2}$, i.e., $W_Q$, $W_K$ and $W_V$ are $l_1$ penalized over the input channels with different strength. 
{\hfill$\square$}\\

{\it Remark 1.} The analysis of Theorem \ref{thm:1} is also capable of combining the ReLU + Layernorm or Batchnorm + ReLU + MLP structures. In some types of transformer models, the tokens will first go through Layernorm or be projected via MLP before entering the self-attention. Via the similar analysis, we can also show the adaptive implicit $l_1$ regularization in these two settings.

\subsection{Tokens Diversification}
Next, we demonstrate the scaled ReLU's token diversification ability by consine similarity.  Following~\cite{gong2021improve} the consine similarity metric is defined as:
\begin{equation}
    \textrm{CosSim(B)} = \frac{1}{n(n-1)}\sum_{i \neq j}\frac{B_{i}B_{j}^{T}}{\lVert B_{i}\rVert\lVert B_{j}\rVert},\label{eq:8}
\end{equation}
where $B_{i}$ represents the $i$-th row of matrix $B$ and $\|\cdot\|$ denotes the $l_2$ norm. Note that if we can ensure $\|B_i\|>b_{\min}$ for $i=1,2,...,n$, the $\textrm{CosSim}(B)$ will in turn be upper bounded by 
\begin{align}
\textrm{CosSim(B)} &\le  \frac{1}{n(n-1)b_{\min}^2}\sum_{i\ne j}B_iB_j^T\notag\\
&= \frac{1}{n(n-1)b_{\min}^2}\left[e^TBB^Te-\sum_{i}\|B_i\|_2^2\right]\notag\\
&\le \frac{1}{n-1}\left(\frac{\|B\|_{op}^2}{b_{\min}^2}-1\right)\label{eq:final_1},
\end{align} 
where $\|\cdot\|_{op}$ denotes matrix operator norm. Based on \eqref{eq:final_1}, as long as $\|B\|_{op}$ and $b_{\min}$ change at the same order, the consine similarity may decease. 
In the following Theorem \ref{theorem2}, we analyze the order of $\|B\|_{op}$ and $\min_i \|B_i\|$.

\begin{theorem}\label{theorem2}
Let $\mathcal{D}$ be a zero mean probability distribution and matrix $A\in\mathcal{R}^{n\times d}$ be a matrix filled whose elements are drawn independently from $\mathcal{D}$ and $\|A\|_{op}\le R\sqrt{nd}$ with $R>0$. Furthermore, we denote $B = ReLU_{\alpha,\beta}(A)$, $\mu_{B} = \mathbbm{E}[B_{i,j}]$ and $\sigma_{B}^2 = \textrm{Var}[B_{i,j}]$ for all $i,j$. Then for $\delta>0$ and $\gamma\in(0,c_0)$ , with probability $1-\delta-2\exp(-cc_0^{-2}\gamma^2d+\log n)$, we have
\begin{align}
\|B\|_{op}\le O\left(\mu\log\left(\frac{1}{\delta}\right)+\sigma\sqrt{\log\left(\frac{1}{\delta}\right)}\right)\notag
\end{align}
and
\begin{align}
    \min_{i}\|B_i\|_2\ge O\left(\sqrt{\mu^2+(1-\gamma)\sigma^2}\right) \notag,
\end{align}
where $c,c_0$ are positive constants, $O(\cdot)$ suppresses the dependence in $n,d$ and $R$.
\end{theorem}

The above result shows that the operator norm and $l_2$ norm for each row of the token matrix after scaled ReLU is proportional to its element-wise mean and standard deviation.  Given the identity transformation (i.e.,  $B = A$) is a special case of the scaled ReLU, matrix $A$ (token matrix before scaled ReLU) enjoys the similar properties. As the ReLU truncates the negative parts of its input, one has $\mu_{B}\ge \mu_{A}$. If we could maintain the same variance level in $B$ and $A$, both $\min_i\|B_i\|_2$ and $\|B\|_{op}$ 
change at order of $O(\mu+\sigma)$ and according to inequality \eqref{eq:final_1}, the cosine similarity becomes smaller from $A$ to $B$.



\textit{Proof of Theorem~\ref{theorem2}:} 

{\bf Upper Bound for $\|B\|_{op}$}.  
Denote $E\in\mathbbm{R}^{n\times d}$ as the matrix filled with 1 and $X = B-\mu E$. We have  $\mathbbm{E}[X] = 0$, $\|X\|_{op}\le  (\beta R+ \beta\alpha+\mu)\sqrt{nd}$ almost surely. Via the matrix Bernstein inequality (e.g., Theorem 1.6  in \citealt{tropp2012user}),
\begin{align}
\mathbbm{P}\left(\left\|X\right\|_{op}\ge t\right)\le (n+d)\exp\left(\frac{-t^2/2}{\sigma_{\max}^2+R_{\max}t/3}\right),\label{eq:new_1}
\end{align}
where 
\begin{align}
    \sigma_{\max}^2 &= \max\{\|\mathbbm{E}[XX^T]\|_{op},\|\mathbbm{E}[X^TX]\|_{op}\} \notag\\
    &=\max\{n\sigma^2,d\sigma^2\}\le (n+d)\sigma^2\notag\\
    R_{\max} &\ge \|X\|_{op}= (\beta R+\beta\alpha+\mu)\sqrt{nd}\notag.
\end{align}
By setting $\delta = (n+d)\exp\left(\frac{-t^2/2}{\sigma_{\max}^2+R_{\max}t/3}\right)$, we can represent $t$ by $\delta$ as:
\begin{align}
    t &= \frac{1}{3}R_{\max}\log\left(\frac{n+d}{\delta}\right)\notag\\
   &+\sqrt{\frac{1}{9}R_{\max}^2\log^2\left(\frac{n+d}{\delta}\right)+2\sigma_{\max}^2\log\left(\frac{n+d}{\delta}\right)}\notag\\
   &\le \frac{2}{3}R_{\max}\log\left(\frac{n+d}{\delta}\right)+\sqrt{2\sigma_{\max}^2\log\left(\frac{n+d}{\delta}\right)},\notag
\end{align}
where last inequality uses the fact that $\sqrt{a+b}\le \sqrt{|a|}+\sqrt{|b|}$.

Then inequality \eqref{eq:new_1} implies the following result holds with probability $1-\delta$:
\begin{align}
   \left\|X\right\|_{op}
   &\le \frac{2}{3}R_{\max}\log\left(\frac{n+d}{\delta}\right)+\sqrt{2\sigma_{\max}^2\log\left(\frac{n+d}{\delta}\right)}\label{eq:new_2}.
\end{align}
Next, combine \eqref{eq:new_2} with the facts $\|B\|_{op}-\|\mu E\|_{op}\le \|X\|_{op}$ and $\|\mu E\|_{op} = \mu\sqrt{nd}$, one has
\begin{align}
\|B\|_{op}&\le O\left(\mu\log\left(\frac{1}{\delta}\right) +\sigma\sqrt{\log\left(\frac{1}{\delta}\right)}\right)\notag,
\end{align}
where we ignore the dependence in $n,d$ and $R$.\\

{\bf Lower Bound for $\|B_i\|$}. Next, we derive the bound for $\|B_i\|$. Since $\|A\|_{op}$ is upper bounded, there exists a constant $c_0$ such that $B_{ij}^2-\mu^2-\sigma^2$ being  centered $c_0\sigma^2$ sub-exponential random variable. Then we are able to apply the Corollary 5.17 in \citealt{vershynin2010introduction}, there exists $c>0$, for $\eta>0$:
\begin{align}
   &\mathbbm{P}\left(\left|\sum_{j}^dB_{ij}^2-d(\mu^2+\sigma^2)\right|\ge \eta d \right)\notag\\
   &\quad \le 2\exp\left(-c\min\left\{\frac{\eta^2}{c_0^2\sigma^4},\frac{\eta}{c_0\sigma^2}\right\}d\right)\notag.
\end{align}
We then set $\eta = \gamma \sigma^2$ for some $\gamma\in(0,c_0)$ such that $   \mu^2+(1-\gamma)\sigma^2>0$. Combining $\|B_i\|^2 = \sum_{j}^dB_{ij}^2$ with above inequality, we have
\begin{align}
   &\mathbbm{P}\left(\|B_i\|\le \sqrt{d(\mu^2+(1-\gamma)\sigma^2)} \right) \le 2\exp\left(-c\gamma^2c_0^{-2}d\right)\notag. 
\end{align}
Therefore via union bound, we have
$$\min_{i}\|B_{i}\|\ge \sqrt{d(\mu^2+(1-\gamma)\sigma^2)} = O(\sqrt{\mu^2+(1-\gamma)\sigma^2})$$ holds with probability $1-2\exp(-c\gamma^2c_0^{-2}d+\log n)$.

{\hfill $\square$}

\section{Experiments}

\begin{table}[ht]
\centering
\begin{tabular}{c|c|c|c|c} 
\hline
                   model  & lr & optimizer &  wm-epoch & Top-1 acc \\
\hline
\multirow{6}{*}{\makecell[c]{ViT$_{p}$\\DeiT-Small}} & 5e-4 & AdamW & 5 & 79.8  \\ 
\cline{2-5}
                  & 1e-3 & AdamW & 5 &  \cellcolor{Light}crash \\ 
\cline{2-5}
                  & 1e-3 & AdamW & 20 &  80.0 \\ 
\cline{2-5}
                  & 5e-4 & SAM & 5 &  79.9 \\ 
\cline{2-5}
                  & 1e-3 & SAM & 5 &  79.6 \\ 
\cline{2-5}
                  & 1e-4 & SAM & 5 &  77.8 \\ 
\hline
\multirow{6}{*}{\makecell[c]{ViT$_{c}$\\DeiT-Small}} & 5e-4 & AdamW  & 5 &  81.6   \\ 
\cline{2-5}
                  & 1e-3 & AdamW & 5 &  \textbf{81.9} \\
\cline{2-5}
                  & 1e-3 & AdamW & 20 &  81.7 \\
\cline{2-5}
                  & 1e-3 & AdamW & 0 &  \cellcolor{Light}crash \\
\cline{2-5}
                  & 5e-4 & SAM & 5 &  81.5 \\ 
\cline{2-5}
                  & 1e-3 & SAM & 5 &  81.7 \\
\cline{2-5}
                  & 1e-4 & SAM & 5 &  79.1 \\ 
\hline
\end{tabular}
\caption{The effects of \textit{conv-stem} using different learning rate (lr), optimizer, warmup epoch (wm-epoch).}\label{conv-stem}
\end{table}

In this section, we conduct extensive experiments to verify the effects of \textit{conv-stem} and scaled ReLU. The ImageNet-1k~\cite{russakovsky2015imagenet} is adopted for standard training and validation. It contains 1.3 million images in the training set and 50K images in the validation set, covering 1000 object classes. The images are cropped to 224$\times$224.

\subsection{The effects of \textit{conv-stem}}

We take DeiT-Small~\cite{touvron2020training} as our baseline, and replace the \textit{patchify-stem} with \textit{conv-stem}. The batchsize is 1024 for 8 GPUs, and the results are as shown in Table~\ref{conv-stem}. From the Table we can see that \textit{conv-stem} based model is capable with more volatile training environment: with \textit{patchify-stem}, ViT$_{p}$ can not support larger learning rate (1e-3) using AdamW optimizer but only works by using SAM optimizer, which reflects ViT$_{p}$ is sensitive to learning rate and optimizer. By adding the \textit{conv-stem}, ViT$_{c}$ can support larger learning rate using both AdamW and SAM optimizers. Interestingly, ViT$_{c}$ achieves 81.9 top-1 accuracy using lr=1e-3 and AdamW optimizer, which is 2.1 point higher than baseline. With \textit{conv-stem}, SAM is no longer more powerful than AdamW, which is a different conclusion as in~\cite{chen2021vision}. After adding \textit{conv-stem}, it still needs warmup, but 5 epochs are enough and longer warmup training does not bring any benefit.

\vspace{-1mm}

\begin{table*}[ht]
\centering
\begin{tabular}{c|c|c|c|c|c|c} 
\hline
                 model & lr & optimizer & wm-epoch & components in \textit{conv-stem} & stride & Top-1 acc  \\ 
\hline
\multirow{17}{*}{{DeiT-Small}} & 1e-3 & AdamW & 5 & 3Conv+3BN+3ReLU+1Proj & (2,1,1,8) & \textbf{81.9} \\ 
\cline{2-7}
                  & 1e-3 & AdamW & 5 & 3Conv+3BN+1Proj & (2,1,1,8) & \cellcolor{Light}crash \\ 
\cline{2-7}
                  & 1e-3 & AdamW & 5 & 3Conv+3ReLU+1Proj & (2,1,1,8) & 81.5 \\
\cline{2-7}
                  & 1e-3 & AdamW & 5 & 3Conv+1Proj &  (2,1,1,8) & \cellcolor{Light}crash \\
\cline{2-7}
                  & 1e-3 & AdamW & 20 & 3Conv+1Proj &  (2,1,1,8) & 80.0 \\
\cline{2-7}
                  & 1e-3 & AdamW & 5 & 3Conv+1Proj+1ReLU & (2,1,1,8) & 79.9 \\
\cline{2-7}
                  & 1e-3 & AdamW & 5 & 1Proj+1BN+1ReLU & (16) & 79.8 \\
\cline{2-7}
                  & 1e-3 & AdamW & 5 & 1Proj+1ReLU & (16) & 79.5 \\
\cline{2-7}
                  & 1e-3 & AdamW & 5 & 1Proj & (16) & \cellcolor{Light}crash \\
\cline{2-7}
                  & 5e-4 & AdamW & 5 & 1Proj (baseline) & (16) & 79.8 \\
                  
\cline{2-7}
                  & 1e-3 & SAM & 5 & 3Conv+3BN+3ReLU+1Proj & (2,1,1,8) & 81.7 \\
\cline{2-7}
                  & 1e-3 & SAM & 5 & 3Conv+3BN+1Proj & (2,1,1,8) & 80.2 \\
\cline{2-7}
                  & 1e-3 & SAM & 5 & 3Conv+3ReLU+1Proj & (2,1,1,8) & 80.6 \\
\cline{2-7}
                  & 1e-3 & SAM & 5 & 3Conv+1Proj &  (2,1,1,8) & \cellcolor{Light}crash \\
\cline{2-7}
                  & 1e-3 & SAM & 20 & 3Conv+1Proj & (2,1,1,8) &  80.4\\
\cline{2-7}
                  & 1e-3 & SAM & 5 & 3Conv+1Proj+1ReLU & (2,1,1,8) & 80.3 \\
\hline
\hline
\multirow{7}{*}{\makecell[c]{DINO-S/16\\100 epoch }} & 5e-4 & AdamW & 10 & 3Conv+3BN+3ReLU+1Proj & (2,1,1,8) & 76.0 \\ 
\cline{2-7}
                  & 5e-4 & AdamW & 10 & 3Conv+3BN+1Proj & (2,1,1,8) & 73.4 \\ 
\cline{2-7}
                   & 5e-4 & AdamW & 10 & 3Conv+3ReLU+1Proj & (2,1,1,8) & 74.8 \\
\cline{2-7}
                   & 5e-4 & AdamW & 10 & 3Conv+1Proj & (2,1,1,8) & 74.1 \\
\cline{2-7}
                   & 5e-4 & AdamW & 10 & 1Proj+1ReLU & (16) & 73.6 \\
\cline{2-7}
                   & 5e-4 & AdamW & 10 & 1Proj+1BN+1ReLU & (16) & 73.3 \\
\cline{2-7}
                   & 5e-4 & AdamW & 10 & 1Proj (baseline) & (16) & 73.6 \\
\hline
\hline
\multirow{7}{*}{VOLO-d1-224} & 1.6e-3 & AdamW & 20 & 3Conv+3BN+3ReLU+1Proj & (2,1,1,4) & \textbf{84.1} \\ 
\cline{2-7}
                  & 1.6e-3 & AdamW & 20 & 3Conv+3BN+1Proj & (2,1,1,4) & 83.6 \\ 
\cline{2-7}
                  & 1.6e-3 & AdamW & 20 & 3Conv+3ReLU+1Proj & (2,1,1,4) & 84.0 \\
\cline{2-7}
                  & 1.6e-3 & AdamW & 20 & 3Conv+1Proj & (2,1,1,4) & \cellcolor{Light}crash \\
\cline{2-7}
                  & 1.6e-3 & AdamW & 20 & 1Proj & 8 & 83.4 \\
\cline{2-7}
                  & 1.6e-3 & AdamW & 20 & 1Proj+1ReLU& 8 & 83.4 \\
\cline{2-7}
                  & 1.6e-3 & AdamW & 20 & 1Proj+1BN+1ReLU& 8 & 83.5 \\
\hline
\end{tabular}
\caption{The effects of scaled ReLU under different settings using three methods. }\label{relu}
\end{table*}

\subsection{The effects of scaled ReLU in \textit{conv-stem}}
We adopt three vision transformer architectures, including both supervised and self-supervised methods, to evaluate the value of scaled ReLU for training ViTs, namely, DeiT~\cite{touvron2020training}, DINO~\cite{caron2021emerging} and VOLO~\cite{yuan2021volo}. For DeiT and VOLO, we follow the official implementation and training settings, only modifying the parameters listed in the head of Table~\ref{relu}; for DINO, we follow the training settings for 100 epoch and show the linear evaluation results as top-1 accuracy.  The results are shown in Table~\ref{relu}. From the Table we can see that scaled ReLU (BN+ReLU) plays a very important role for both stable training and boosting performance. Specifically, without ReLU the training will be crashed under 5 warmup epoch in most cases, for both AdamW and SAM optimizers; increasing warmup epoch will increase the stabilization of training with slightly better results; with scaled ReLU, it can boost the final performance largely in stable training mode. The full \textit{conv-stem} boosts the performance of DeiT-Small largely, 2.1 percent compared with the baseline, but by removing ReLU or scaled ReLU the performance will decrease largely; the same trend holds for both DINO and VOLO. For the \textit{patchify-stem}, after adding ReLU or scaled ReLU it can stabilize the training by supporting a large learning rate. In addition, scaled ReLU has faster convergence speed. For DeiT-Small, the top-1 accuracy is 18.1 vs 10.6 at 5 epoch, 53.6 vs 46.8 at 20 epoch, 63.8 vs 60.9 at 50 epoch, for conv-stem and patchify-stem, respectively.

\vspace{-2mm}

\begin{table}[ht]
\centering
\begin{tabular}{c|c|c} 
\hline
components in \textit{conv-stem} & stride &  top-1 acc \\ 
\hline
 3Conv+3BN+3ReLU+1Proj & (2,1,1,8) &  \textbf{81.9} \\ 
\hline
 3Conv+3BN+3ReLU+1Proj & (2,2,2,2) &  81.0 \\  
\hline
 3Conv+1Proj &  (2,1,1,8) & \cellcolor{Light}crash \\
\hline
  3Conv+1Proj &  (2,2,2,2) & \cellcolor{Light}crash \\
 \hline
3Conv+1Proj (wm-epoch=20) &  (2,1,1,8) & 80.0 \\
\hline
3Conv+1Proj (wm-epoch=20) &  (2,2,2,2) & 79.7 \\
\hline
 3Conv+1Proj+1ReLU & (2,1,1,8) & 79.9   \\
 \hline
 3Conv+1Proj+1ReLU & (2,2,2,2) & 79.9   \\
\hline
\end{tabular}
\caption{The effects of stride in \textit{conv-stem} for DeiT-Small.}\label{stride}
\end{table}

\subsection{Scaled ReLU diversifies tokens}
To analyze the property of scaled ReLU diversifying tokens, we adopt the quantitative metric
\textbf{layer-wise cosine similarity between tokens} as defined in formula~\ref{eq:8}.  

\begin{figure}[ht]
  \begin{minipage}{0.98\linewidth}
    \begin{center}
      \includegraphics[width=\linewidth]{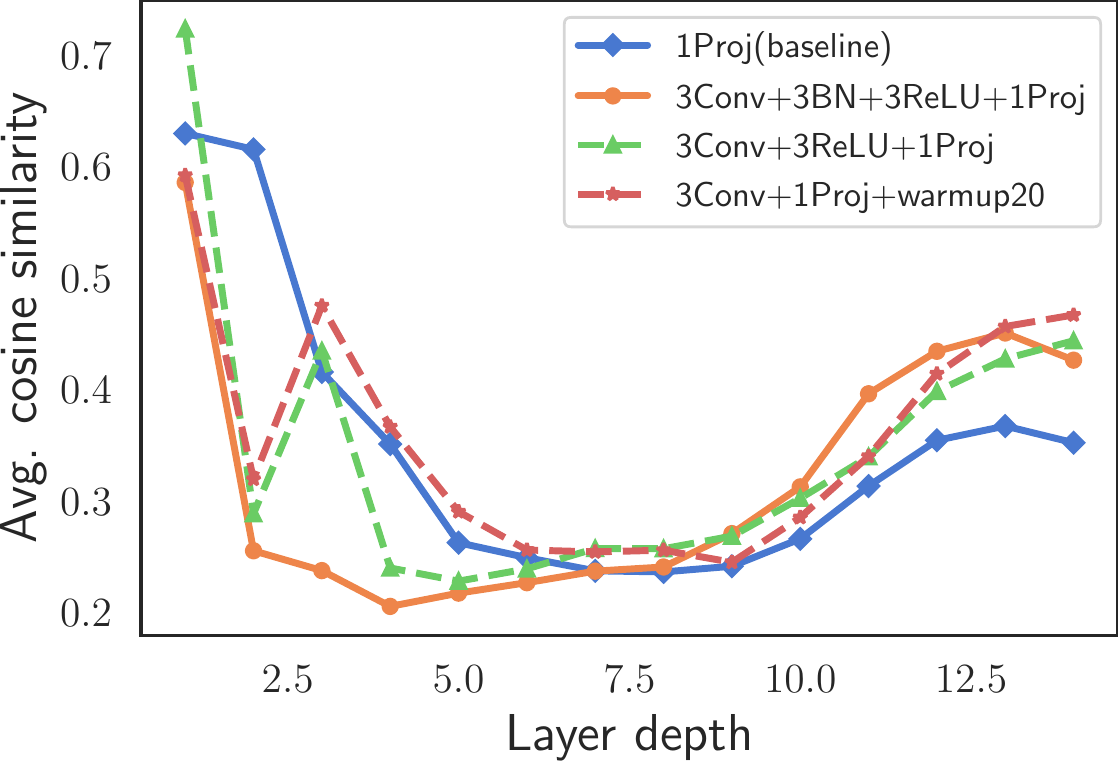}
    \end{center}
  \end{minipage}
  \caption{Layer-wise cosine similarity of tokens for DeiT-Small. }\label{properties}
\end{figure}

We regard the \textit{conv-stem} as one layer and position embedding as another layer in the \textit{ViT-stem}, thus the total layers of ViT$_{c}$ is 14 (plus 12 transformer encoder layers). The layer-wise cosine similarity of tokens are shown in Figure~\ref{properties}. From the Figure we can see that position embedding can largely diversify the tokens due to its specific position encoding for each token. Compared with baseline (1Proj)~\cite{touvron2020training}, the full \textit{conv-stem} (3Conv+3BN+3ReLU+1Proj) can significantly diversify the tokens at the lower layers to learn better feature representation, and converge better at higher layers for task-specific feature learning. Interestingly, 3Conv+3ReLU+1Proj and 3Conv+1Proj+warmup20 have the similar trend which reflects that ReLU can stabilize the training as longer warmup epochs.

\vspace{-3mm}

\vspace{-3mm}

\subsection{The effects of stride in \textit{conv-stem}}
According to the work~\cite{xiao2021early}, the stride in the \textit{conv-stem} matters for the final performance. We also investigate this problem in the context of VOLO \textit{conv-stem} for DeiT-Small. We keep the kernel size unchanged, and only adjust the stride and its corresponding padding. The default warmup epoch is 5 unless otherwise noted. The results are shown in Table~\ref{stride}. From this Table we can see that the average stride (2,2,2,2) is not better than (2,1,1,8), and it can not stabilize the training either.

\vspace{-1mm}

\subsection{Transfer Learning: Object ReID}
In this section, we transfer the DINO-S/16 (100 epoch) on ImageNet-1k to object ReID to further demonstrate the effects of \textit{conv-stem}. We fine-tune the DINO-S/16 shown in Table \ref{relu} on Market1501 \cite{Market1501} and MSMT17 \cite{MSMT17} datasets. We follow the baseline \cite{he2021transreid} and follow the standard evaluation protocol to report the Mean Average Precision (mAP) and Rank-1 accuracies. All models are trained with the baseline learning rate (1.6e-3) and a larger learning rate (5e-2). The results are shown in Table \ref{reid}. From the Table we can see that the full \textit{conv-stem} not only achieves the best performance but also supports both the large learning rate and small learning rate training. Without ReLU or BN+ReLU, in most cases, the finetuning with a large learning rate will crash. Interestingly, the finetuning with DINO is sensitive to the learning rate, a smaller learning rate will achieve better performance.

\vspace{-1mm}

\subsection{Scaled ReLU/GELU in Transformer Encoder}
In transformer encoder layer, the feed-forward layers (ffn) adopt LayerNorm+GELU block, and in this section, we investigate this design using DeiT-Small and VOLO-d1-224, using the training parameters for the best performance in Table~\ref{relu}. The motivation to investigate ReLU and GELU is to show whether GELU is better than ReLU for \textit{conv-stem} design, as GELU achieves better results than ReLU for transformer encoder. We first remove the LayerNorm layer in ffn, the training directly crashes in the first few epochs. And then, we replace the GELU with RELU, the performance drops largely, which reflects that GELU is better than ReLU for ffn. Next, we replace ReLU with GELU in \textit{conv-stem}, the performance drops a little bit, demonstrating that ReLU is better than GELU for \textit{conv-stem}. Lastly, we rewrite the MLP implementation in ffn by replacing the fc+act block with Conv1D+BN1D+GELU (Conv1D equals to fc, and the full implementation is shown in supplemental material of Algorithm 2), and the performance drops, especially for VOLO. It might confirm the conclusion in NFNet~\cite{brock2021high} that batch normalization constrains the extreme performance, making the network sub-optimal. All the results are shown in Table~\ref{encoder}. 

\begin{table}[ht]
\centering
\small
\setlength{\tabcolsep}{4pt}
\begin{tabular}{c|c|cc|cc} 
\hline
 & & \multicolumn{2}{c|}{Market1501} & \multicolumn{2}{c}{MSMT17} \\
 components in \textit{conv-stem} & lr & mAP & R-1 &mAP & R-1 \\
 \hline
 3Conv+3BN+3ReLU+1Proj & \multirow{7}{*}{1.6e-3} &\textbf{84.3} &\textbf{93.5}&\textbf{56.3} &\textbf{78.7} \\
 3Conv+3BN+1Proj &  &83.6 &92.9 &55.1 &77.8 \\
 3Conv+3ReLU+1Proj & &81.7 &91.9  &51.5 &75.2 \\
 3Conv+1Proj &  &83.0 &92.7  &52.1 &74.0 \\
 1Proj+1ReLU&  & 84.2 &93.1 &53.6 &75.5 \\
 1Proj+1BN+1ReLU &  &84.1 &92.8 &55.7 &77.5 \\
 1Proj (baseline) & & 84.1 &93.1 &54.9 &76.8 \\
 \hline  
 3Conv+3BN+3ReLU+1Proj & \multirow{7}{*}{5e-2} &76.8 &89.7 &48.5 &72.1 \\
 3Conv+3BN+1Proj &  & \multicolumn{4}{c}{\cellcolor{Light}crash}  \\
 3Conv+3ReLU+1Proj &  &\multicolumn{4}{c}{\cellcolor{Light}crash} \\
 3Conv+1Proj &  & \multicolumn{4}{c}{\cellcolor{Light}crash}\\ 
 1Proj+1ReLU &  &69.5 & 86.1 & 36.1 & 36.0 \\
 1Proj+1BN+1ReLU &  & 77.6 & 90.6 & 46.2 & 88.6 \\ 
 1Proj (baseline) &  & \multicolumn{4}{c}{\cellcolor{Light}crash}\\ 
  \hline
\end{tabular}
\caption{The comparisons with different components in \textit{conv-stem} based on DINO for finetuning ReID tasks. }\label{reid}
\end{table}

\vspace{-3mm}

\subsection{Self-supervised + supervised training}
To further investigate the training of ViTs, we adopt the DINO self-supervised pretrained ViT-Small model~\cite{caron2021emerging} on ImageNet-1k and use it to initialize the ViT-Small model to finetune on ImageNet-1k using full labels. The results are shown in Table~\ref{twostagetraining}. From this Table we can see that using a self-supervised pretrained model for initialization, ViT$_{p}$ achieve 81.6 top-1 accuracy using SAM optimizer, which is 1.8 percent point higher than baseline. However, according to the analysis in~\cite{newell2020useful}, with large labelled training data like Imagenet-1k dataset, the two stage training strategy will not contribute much (below 0.5 percent point). By adding \textit{conv-stem}, the peak performance of ViT$_{c}$ can reach 81.9 which is higher than two stage training, which reflects that previous ViTs models are far from being well trained. 

\begin{table}[ht]
\centering
\begin{tabular}{c|c|c} 
\hline
                 model & design  & Top-1 acc  \\ 
\hline
\multirow{5}{*}{\makecell[c]{DeiT-Small$_{c}$}} 
                  & LayerNorm removed in ffn & {\cellcolor{Light}crash} \\ 
\cline{2-3}
                  & GELU$\rightarrow$ReLU in ffn & 80.3(1.6$\downarrow$) \\  
\cline{2-3}
                  & ReLU$\rightarrow$GELU in \textit{conv-stem} & 81.7(0.2$\downarrow$) \\
\cline{2-3}
                  & MLP$\rightarrow$Conv1D+BN+GELU & 81.7(0.2$\downarrow$) \\
\cline{2-3}
                  & MLP$\rightarrow$Conv1D+GELU & 82.0(0.1$\uparrow$) \\
\hline
\multirow{5}{*}{\makecell[c]{VOLO-d1$_{c}$}} 
                  & LayerNorm removed in ffn & {\cellcolor{Light}crash} \\ 
\cline{2-3}
                  & GELU$\rightarrow$ReLU in ffn & 83.5(0.6$\downarrow$) \\ 
\cline{2-3}
                  & ReLU$\rightarrow$GELU in \textit{conv-stem} & 84.0(0.1$\downarrow$) \\
\cline{2-3}
                  & MLP$\rightarrow$Conv1D+BN+GELU & 83.2(0.9$\downarrow$) \\
\cline{2-3}
                  & MLP$\rightarrow$Conv1D+GELU & 84.0(0.1$\downarrow$) \\
\hline
\end{tabular}
\caption{The comparisons among different designs using scaled ReLU/GELU. }\label{encoder}
\end{table}

\begin{table}
\centering
\begin{tabular}{c|c|c|c|c} 
\hline
                 model & lr & optimizer & wm-epoch  & Top-1 acc  \\ 
\hline
\multirow{6}{*}{\makecell[c]{DeiT-Small$_{p}$ \\ TST }} & 1e-4 & AdamW & 5 & 81.2 \\ 
\cline{2-5}
                  & 5e-4 & AdamW & 5 & 81.3 \\ 
\cline{2-5}
                  & 1e-3 & AdamW & 5  & 80.1 \\
\cline{2-5}
                  & 1e-4 & SAM & 5 & 81.6 \\ 
\cline{2-5}
                  & 5e-4 & SAM & 5 & 81.1 \\ 
\cline{2-5}
                  & 1e-3 & SAM & 5  & 80.1 \\
\hline
\multirow{2}{*}{\makecell[c]{DeiT-Small$_{c}$\\ OST }} & 1e-3 & AdamW & 5 & 81.9 \\ 
\cline{2-5}
                  & 1e-3 & SAM & 5 & 81.7 \\ 
\hline
\end{tabular}
\caption{The comparisons between two-stage training (TST, self-supervised + supervised training) and only supervised training (OST) on ImageNet-1k. }\label{twostagetraining}
\end{table}

\vspace{-1mm}

\subsection{Scaled Dataset Training}
In order to verify that the previous ViTs are not well trained, we adopt the DINO pretrained ViT-Small model~\cite{caron2021emerging} on ImageNet-1k to initialize the ViT-Small model, and finetune on ImageNet-1k using portion of full labels, containing 1000 classes. We adopt the original~\textit{patchify-stem} and SAM optimizer for this investigation.  The results are shown in Table~\ref{datasize}. It can be seen that even using self-supervised pretrained model for initialization, using only 10\% of ImageNet-1k data for training, it only achieves 67.8\% accuracy, much worse than the linear classification accuracy using full data (77.0\%)~\cite{caron2021emerging}. With the data-size increasing, the performance improves obviously, and we do not see any saturation in the data-size side. This performance partly demonstrates that ViT is powerful in fitting data and current ViT models trained on ImageNet-1k is not trained enough for its extreme performance. 

\begin{table}[ht]
\centering
\begin{tabular}{c|c|c|c|c} 
\hline
                 model & lr & optimizer & data-size  & Top-1 acc  \\ 
\hline
\multirow{10}{*}{\makecell[c]{DeiT-Small$_{p}$ \\ TST }} 
                  & 1e-4 & SAM & 10\% & 67.8 \\ 
\cline{2-5}
                  & 1e-4 & SAM & 20\% & 73.5 \\ 
\cline{2-5} 
                  & 1e-4 & SAM & 30\% & 76.0 \\ 
\cline{2-5}
                  & 1e-4 & SAM & 40\% & 77.6 \\ 
\cline{2-5}
                  & 1e-4 & SAM & 50\% & 79.0 \\ 
\cline{2-5}
                 & 1e-4 & SAM & 60\% & 79.8 \\ 
\cline{2-5}
                 & 1e-4 & SAM & 70\% & 80.4 \\ 
\cline{2-5}
                 & 1e-4 & SAM & 80\% & 80.9 \\ 
\cline{2-5}
                 & 1e-4 & SAM & 90\% & 81.4 \\ 
\cline{2-5}
                 & 1e-4 & SAM & 100\% & 81.6 \\ 
\cline{2-5}
\hline
\end{tabular}
\caption{The comparisons among different portion of ImageNet-1k for two-stage training (TST, self-supervised + supervised training) training. }\label{datasize}
\end{table}

\vspace{-3mm}

\section{Conclusion}
In this paper, we investigate the training of ViTs in the context of \textit{conv-stem}. We theoretically and empirically verify that the scaled ReLU in the \textit{conv-stem} matters for robust ViTs training. It can stabilize the training and improve the token diversity for better feature learning. Using \textit{conv-stem}, ViTs enjoy a peak performance boost and are insensitive to the training parameters. Extensive experiments unveil the merits of \textit{conv-stem} and demonstrate that previous ViTs are not well trained even if they obtain better results in many cases compared with CNNs.

{\small
\bibliography{aaai22}
}

\section{Appendix}
\definecolor{codegreen}{rgb}{0,0.6,0}
\definecolor{codegray}{rgb}{0.5,0.5,0.5}
\definecolor{codepurple}{rgb}{0.58,0,0.82}
\definecolor{backcolour}{rgb}{255,255,255}

\lstdefinestyle{mystyle}{
  backgroundcolor=\color{backcolour},   commentstyle=\color{codegreen},
  keywordstyle=\color{magenta},
  numberstyle=\tiny\color{codegray},
  stringstyle=\color{codepurple},
  basicstyle=\ttfamily\scriptsize,
  breakatwhitespace=false,         
  breaklines=true,                 
  captionpos=b,                    
  keepspaces=true,                 
  numbers=left,                    
  numbersep=8pt,                  
  showspaces=false,                
  showstringspaces=false,
  showtabs=false,                  
  tabsize=5
}

\lstset{style=mystyle}
\onecolumn

\begin{algorithm*}[ht]
\caption{Source codes of Patch Embedding for \textit{conv-stem}.}\label{conv_stem}
\begin{lstlisting}[language=Python]
class PatchEmbed(nn.Module):
    def __init__(self,img_size=224,conv_stem=True,stem_stride=2,patch_size=16,in_chans=3,hid_dim=64,embed_dim=384):
        super().__init__()
        self.conv_stem = conv_stem
        if conv_stem:
            self.conv = nn.Sequential(
                nn.Conv2d(in_chans,hid_dim,kernel_size=7,stride=stem_stride,padding=3,bias=False),#112x112
                nn.BatchNorm2d(hid_dim),
                nn.ReLU(inplace=True),
                nn.Conv2d(hid_dim,hid_dim,kernel_size=3,stride=1,padding=1,bias=False),#112x112
                nn.BatchNorm2d(hid_dim),
                nn.ReLU(inplace=True),
                nn.Conv2d(hid_dim,hid_dim,kernel_size=3,stride=1,padding=1,bias=False),#112x112
                nn.BatchNorm2d(hid_dim),
                nn.ReLU(inplace=True))
        self.proj = nn.Conv2d(hid_dim,embed_dim,kernel_size=patch_size//stem_stride,stride=patch_size//stem_stride)
        self.num_patches = (img_size//patch_size)*(img_size//patch_size)

    def forward(self, x):
        if self.conv_stem:
            x = self.conv(x)
        x = self.proj(x) #B,C,H,W
        x = x.flatten(2).transpose(1,2) #B,N,C
        return x
\end{lstlisting}
\end{algorithm*}

\begin{algorithm*}[ht]
\caption{Source codes of two Mlp implementations.}\label{mpl}
\begin{lstlisting}[language=Python]
class Mlp(nn.Module): ## original implementation
    def __init__(self, in_features, hidden_features=None, out_features=None, act_layer=nn.GELU, drop=0.):
        super().__init__()
        out_features = out_features or in_features
        hidden_features = hidden_features or in_features
        self.fc1 = nn.Linear(in_features, hidden_features)
        self.act = act_layer()
        self.fc2 = nn.Linear(hidden_features, out_features)
        self.drop = nn.Dropout(drop)

    def forward(self, x):
        x = self.fc1(x)
        x = self.act(x)
        x = self.drop(x)
        x = self.fc2(x)
        x = self.drop(x)
        return x

class Mlp(nn.Module): ## revised implementation using Conv1D+BN1D+GELU
    def __init__(self, in_features, hidden_features, act_layer=nn.GELU,drop=0.):
        super().__init__()
        self.w1 = nn.Sequential(
                nn.Conv1d(in_features, hidden_features,kernel_size=1,stride=1),
                nn.BatchNorm1d(hidden_features),
                nn.GELU(),
            )

        self.w2 = nn.Sequential(
                nn.Conv1d(hidden_features, in_features,kernel_size=1,stride=1),
                nn.BatchNorm1d(in_features),
                nn.GELU(),
            )
        self.drop = nn.Dropout(drop)

    def forward(self, x):
        x = x.permute(0,2,1)
        x = self.w1(x)
        x = self.drop(x)
        x = self.w2(x)
        x = x.permute(0,2,1)
        return x

\end{lstlisting}
\end{algorithm*}
\end{document}